\documentclass[runningheads]{llncs}
\usepackage[T1]{fontenc}
\usepackage{hyperref}
\usepackage{url}
\usepackage{graphicx}
\usepackage{cite}
\usepackage{picinpar}
\usepackage{amsmath}
\usepackage{url}
\usepackage{flushend}
\usepackage{colortbl}
\usepackage{soul}
\usepackage{multirow}
\usepackage{pifont}
\usepackage{color}
\usepackage{alltt}
\usepackage{enumerate}
\usepackage{siunitx}
\usepackage{breakurl}
\usepackage{epstopdf}
\usepackage{pbox}
\usepackage{amsmath}
\usepackage{amssymb}
\usepackage{mathtools}
\usepackage{amsfonts}
\usepackage{dsfont}
\usepackage[table,xcdraw]{xcolor}
\usepackage{threeparttable}
\usepackage{colortbl}
\usepackage{multirow}
\usepackage{booktabs}
\usepackage{subcaption}
\usepackage{siunitx}
\usepackage[switch]{lineno}
\usepackage[ruled,vlined]{algorithm2e}
\usepackage{algpseudocode}
\definecolor{Cerulean}{rgb}{0,0,0.95}
\definecolor{LimeGreen}{rgb}{0.15,0.65,0.15}
\definecolor{RoyalBlue}{rgb}{0.25,0.41,0.88}
\definecolor{best}{RGB}{255, 223, 186}
\definecolor{best2}{RGB}{204, 229, 255}
\definecolor{Rose}{rgb}{1.0, 0.15, 0.21}
\definecolor{Orange}{rgb}{1.0, 0.5, 0.0}
\definecolor{Gray}{gray}{0.6}
\definecolor{Black}{gray}{0.0}
\definecolor{Purple}{rgb}{0.77,0.12,0.64}
\definecolor{FullBlue}{rgb}{0,0,1}

\definecolor{best}{RGB}{255, 223, 186}
\definecolor{best2}{RGB}{204, 229, 255}

\newcommand{\NickName}{\textit{DrivingWorld}}
\begin{document}
\title{DrivingWorld: Constructing World Model for Autonomous Driving via Video GPT}

\titlerunning{DrivingWorld}
%
\author{Xiaotao Hu\inst{1,2}\thanks{Corresponding author. E-mail: xiaotao.hu@outlook.com} \and
Mingkai Jia\inst{1,2} \and Xiaoyang Guo\inst{2} \and Qian Zhang\inst{2} \and Xiao-xiao Long\inst{3} \and
Wei Yin\inst{2}}
\authorrunning{Hu et al.}
%
\institute{Hong Kong University of Science and Technology, Hong Kong SAR \and
Horizon Robotics, Shanghai, China \and
Nanjing University, Nanjing, China}
\maketitle              
%

\begin{abstract}

Recent successes in autoregressive (AR) generation models, such as ChatGPT, have inspired efforts to extend these techniques to visual tasks. Although various approaches have been proposed to construct video-based world models for autonomous driving, their performance has consistently remained suboptimal. This limitation arises because the classic GPT framework is designed for one-dimensional contextual information and lacks the capacity to model the high-dimensional dynamics essential for video generation.
In this paper, we present \NickName, a framewise autoregressive world model for autonomous driving that effectively captures both spatial and temporal dynamics to predict high-fidelity future states. Our method employs a next-frame prediction strategy to ensure temporal coherence between consecutive frames and a next-token prediction strategy to capture spatial details within each frame. To further enhance generalization, we introduce balanced attention and masking strategies for token prediction, enabling long-term future prediction and precise control.
Experiments demonstrate that our approach produces video clips that are twice as long—up to 40 seconds—while maintaining high fidelity and consistency over previous methods.

\end{abstract}    
\section{Introduction}
\label{intro}

Predicting future events is crucial for autonomous systems, as it enables informed decision-making for safe and efficient navigation in complex environments, while enhancing adaptability to dynamic conditions and improving both performance and reliability. To this end, researchers are exploring world models, which predict future states from past observations and potential actions. Specifically, a world model takes a sequence of past states (observations, actions, and other relevant variables) as input and outputs the predicted state at the next time step. By learning structured representations of the environment, such models support more effective decision-making in autonomous systems.

However, most existing world model approaches for autonomous driving rely heavily on extensive annotations (such as high-definition maps, 3D bounding boxes, and linguistic labels), limiting their scalability. The scarcity and slow growth of publicly available annotated datasets hinder progress, while even state-of-the-art methods remain ineffective in handling out-of-distribution scenarios.
    
Next-token autoregressive (AR) models have achieved remarkable success in Natural Language Processing and image generation, largely owing to their self-supervised training paradigm, which can effectively leverage large-scale unlabeled data. We aim to exploit this advantage to address the data limitations faced by world models in autonomous driving. However, directly adopting a next-token AR framework by discretizing multimodal inputs into a one-dimensional token sequence presents \textbf{two critical limitations}:

\textbf{Control Signal Degradation}: The token composition is severely imbalanced—each temporal state contains merely two control tokens (orientation + location) versus hundreds of image tokens. This imbalance introduces two fundamental issues:
1) \textbf{\textit{Attention Dilution}}: When predicting pose-related tokens, the attention mechanism must process a disproportionate number of image tokens ($> 99$\% of total tokens). The massive visual features inevitably dominate attention weights, diluting the connection with historical control signals~\cite{tiezzi2025back, li2025minimax, gu2023mamba, child2019generating, liu2021swin}.
2) \textbf{\textit{Temporal Disconnection}}: The sparse distribution of control tokens (2 control tokens vs. 500+ image tokens) hinders effective modeling of vehicle dynamics. The traditional attention mechanism struggles to establish coherent temporal dependencies for control signals across frames, as they get intermittently submerged in visual feature surges (see Section III-B for detailed experiments).

\textbf{Computational Inefficiency}: The traditional attention mechanism requires each predicted token to interact with all preceding tokens, resulting in $(\mathcal{O}(T^2n^2))$  computational complexity proportional to the number of historical frames. This exhaustive cross-token interaction imposes substantial computational overhead.
These limitations collectively impair the model's ability to generate accurate vehicle control signals, ultimately compromising the world model's controllability when using pose modalities for trajectory prediction.

The key contributions of this work are threefold:
\begin{itemize}
    \item  We propose \NickName, a novel self-supervised, frame-wise world model for autonomous driving that decouples temporal dynamics from spatial structures in world modeling.
    \item Our framework employs a novel dual-strategy approach: \textbf{Inter-Frame Temporal Modeling} and \textbf{Internal-Frame Spatial Reasoning}. It explicitly separates temporal modeling from cross-modal spatial reasoning while maintaining their interaction, addressing both the temporal coherence and modality imbalance challenges.
    \item Experiments show that the proposed model not only achieves good generalization and generates realistic video sequences over 40 seconds long, but also produces effective policies for vehicle control.
\end{itemize}

\begin{figure*}[t]
    \centering
    \includegraphics[width=0.9\textwidth]{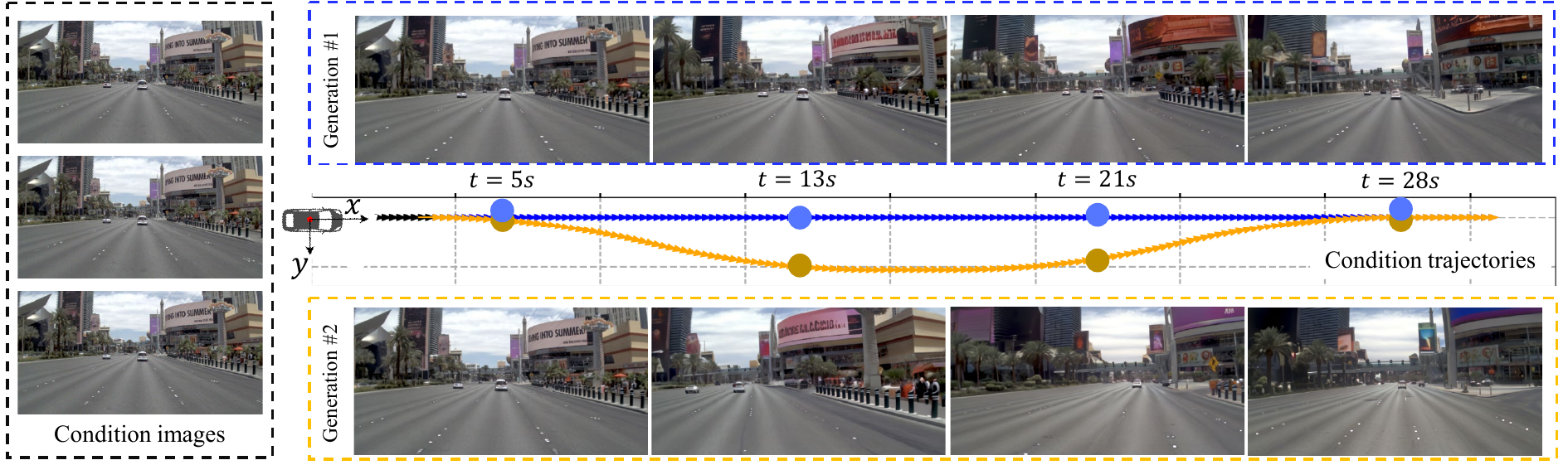}
    \caption{Controllable generation results of our method. Our method takes a short video clip as input and can generate multiple possible future driving scenarios conditioned on different pre-defined trajectory paths. We use one straightforward path and one curved path as examples, and the results show that our method achieves accurate control and future prediction.}
    \label{Fig: first page fig}
\end{figure*}

\section{Related Work}
\label{related}

\subsection{World Model}
World models capture a comprehensive representation of the environment and forecasts future states based on a sequence of actions~\cite{lecun2022path}, extensively explored in both game and laboratory environments.
Dreamer~\cite{hafner2019dream} trained a latent dynamics model using past experiences to forecast state values and actions within a latent space.
Built upon the original Dreamer model, DreamerV2~\cite{hafner2020mastering} achieved human-level performance in Atari games.
DreamerV3~\cite{hafner2023mastering} employed larger networks and successfully learned to acquire diamonds in Minecraft from scratch.

Recently, world models for driving scenarios have garnered significant attention in both academia and industry.
Most previous works have been limited to simulators or well-controlled lab environments. 
Drive-WM~\cite{wang2024driving} explored diffusion-based real-world driving planners. 
GenAD~\cite{yang2024generalized} addressed dynamics in driving scenes with temporal reasoning blocks.
Vista~\cite{gao2024vista} explored video generation for driving scenarios using diffusion models. However, reliance on annotated data in previous approaches severely restricts the applicability of world models. 
GAIA-1~\cite{hu2023gaia} investigated real-world driving planners based on AR models but had large parameters and computational demands. 
In this paper, we propose an efficient, self-supervised world model within an AR framework for autonomous driving scenarios, offering lower computational cost and superior performance.

\begin{figure}[t]
    \centering
    \includegraphics[width=0.9\textwidth]{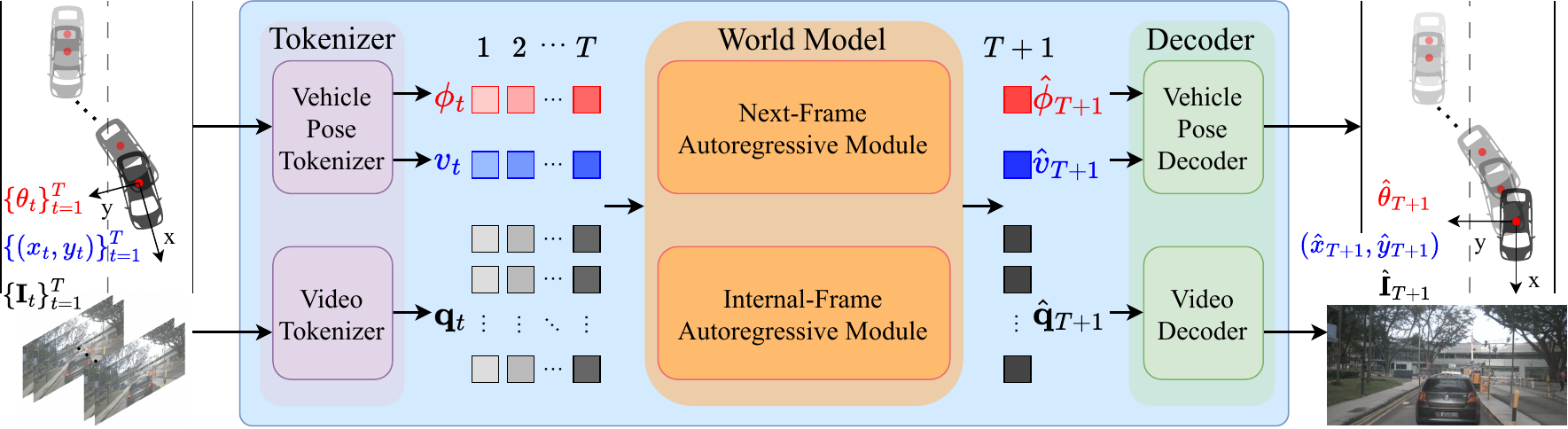}
    \caption{Pipeline of \NickName. The vehicle orientations $\{\theta_{t}\}_{t=1}^{T}$, ego locations $\{(x_{t}, y_{t})\}_{t=1}^{T}$, and a front-view image sequence $ \{\mathbf{I}_{t}\}_{t=1}^{T} $ are taken as the conditional input, tokenized as latent embeddings. Our proposed multimodal world model attempts to comprehend them and forecast the future states, detokenized to the vehicle orientation $\hat{\theta}_{T+1}$, location $(\hat{x}_{T+1}, \hat{y}_{T+1})$, and the front-view image $\hat{\textbf{I}}_{T+1}$. Our AR approach generates 40+ second videos. }
    \label{fig:pipeline}
\end{figure}

\subsection{VQVAE} VQVAE~\cite{van2017neural} learned a discrete codebook representation via vector quantization to model image distributions. 
VQGAN~\cite{esser2021taming} improved realism by incorporating LPIPS loss~\cite{zhang2018unreasonable}.
MoVQ~\cite{zheng2022movq} tackled spatial normalization issue via variant-aware vector quantization.
LlamaGen~\cite{sun2024autoregressive} further fine-tuned VQGAN, showing that a larger codebook size could enhance reconstruction performance. 
However, these methods focused on single-image processing tasks, lacking the temporal modeling capabilities necessary to maintain coherent frame-to-frame consistency in video.
To address this, we propose a temporal-aware tokenizer and decoder.

\subsection{Video Generation}
Currently, there are three mainstream video generation models: GAN-based, diffusion-based, and GPT-based methods. GAN-based methods~\cite{yu2022generating} often suffer from mode collapse, limiting output diversity.
Additionally, the adversarial learning between the generator and discriminator often causes instability.
Diffusion-based methods~\cite{blattmann2023stable} struggle with precise video control due to inherent stochasticity in the denoising process. Maintaining object consistency during denoising is also a significant challenge.
On the other hand, traditional GPT-based methods~\cite{yan2021videogpt} allow for a certain level of control, but their computational cost grows quadratically with the sequence length, significantly impacting model efficiency. In this paper, we propose a novel dual-strategy world model framework, which ensures controllability while significantly reducing computational cost and improving model efficiency.

\section{Method}
\label{method}

Our proposed world model, \NickName, leverages an AR architecture to predict future states with high efficiency, capable of long-horizon predictions, i.e. \textit{beyond $400$ frames}.
Figure~\ref{fig:pipeline} presents our pipeline, which consists of tokenizers for encoding vehicle poses and videos into discrete tokens, the world model for future forecasting, and the decoders to reconstruct vehicle poses and videos from discrete tokens.
Specifically, \NickName~ focuses on predicting the state at time step $T+1$ based on the historical states from time steps $1$ to $T$, where each state consists of a vehicle pose $( \theta_t,  x_t,  y_t)$ and the corresponding frame $\mathbf{I}_t$ in temporal order. The prediction is performed in two steps: (1) the model first predicts the vehicle’s next action $( \hat{\theta}_{T+1},  \hat{x}_{T+1},  \hat{y}_{T+1})$; (2) this predicted action, together with the historical inputs, is then used to generate the next frame $\hat{\mathbf{I}}_{T+1}$.
$\theta_t$ and $(x_t, y_t)$ are the vehicle's orientations and locations, while $\mathbf{I}_t$ represent the video frames. 
Vehicle's poses $(\theta_t, x_t, y_t)$ are tokenized to $\phi_t$ and $v_t$, and $\textbf{I}_t$ is tokenized to $\textbf{q}_t$. 

\subsection{Tokenizer}
\label{sec:tokenizer}
Tokenizer aims to transform continuous inputs to discrete tokens. We explore token discretization for two types of signals, frames $\{\mathbf{I}_{t}\}_{t=1}^{T}$ and vehicle poses $\{\mathbf{(\theta}_{t}, x_t, y_t)\}_{t=1}^{T}$.

\noindent
\textbf{Video Tokenizer.} We follow VQGAN~\cite{esser2021taming} to quantize each image $\mathbf{I}_t$ into a set of tokens $\mathbf{q}_t \in [K]^{H\times W}$. However, VQGAN~\cite{esser2021taming} lacks temporal processing, causing noticeable inconsistency. In our implementation, we insert a causal attention layer both \textit{before} and \textit{after} VQGAN~\cite{esser2021taming} quantization, where the attention operates on the temporal dimension.

\begin{figure}[t]
    \centering
    \includegraphics[width=0.9\textwidth]{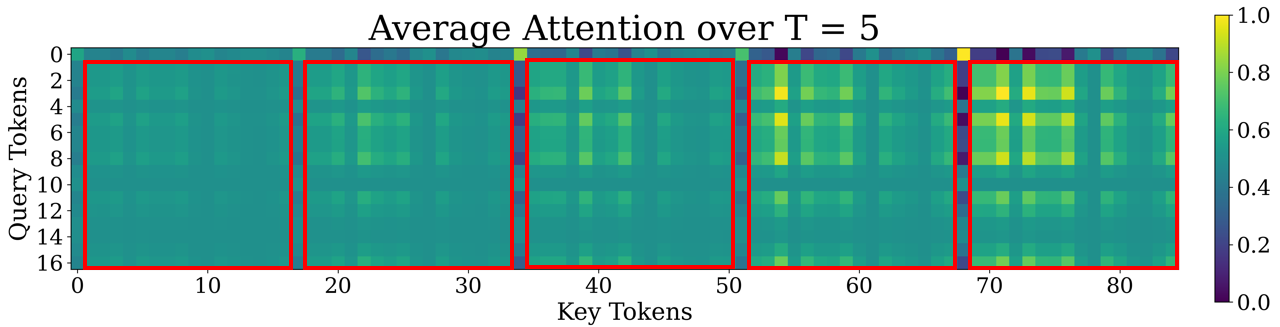}
    \caption{Attention map visualization. Attention map from the fifth Transformer block of a naive AR model~\cite{radford2019language}. Red boxes highlight control signal degradation, where image tokens dominate attention over control tokens. Brighter colors toward the right indicate temporal disconnection, with query tokens favoring nearby timesteps.}
    \label{fig:attn_map_vis}
\end{figure}

\noindent
\textbf{Vehicle Pose Tokenizer.} 
To accurately represent the vehicle's orientation \( \theta \) and location \( (x, y) \), we adopt a coordinate system centered at the ego vehicle, as depicted in Figure~\ref{fig:pipeline}. 
Instead of global poses, we adopt relative poses between adjacent time steps.
This is because global poses present a significant challenge due to the increasing magnitude of global pose values over long-term sequences. As sequences extend, managing global pose values becomes more difficult, complicating normalization and reducing model robustness.

For the sequence of the vehicle's orientation ${\{\theta_t\}}_{t=1}^{T}$ and location ${\{(x_t, y_t)\}}_{t=1}^{T}$, we compute relative values at each time step based on the previous one. The relative location and orientation at the first time step are set to zero. The pose sequence is denoted by ${\{\Delta \theta_t\}}_{t=1}^{T}$ and ${\{(\Delta x_t, \Delta y_t)\}}_{t=1}^{T}$.
To tokenize them, we discretize the pose's surrounding space. 
Specifically, we discretize the orientation into $\alpha$ categories, and the $X$ and $Y$ axes into $\beta$ and $\gamma$ bins, respectively. The relative pose at time $t$ is tokenized as follows:
\begin{equation}
\begin{split}
\phi_t &= \left\lfloor\frac{\Delta \theta_t - \theta_{min}}{\theta_{max} - \theta_{min}}\alpha \right\rfloor, \\ 
v_t &= \left\lfloor\frac{\Delta x_t -  x_{min}}{x_{max} - x_{min}}\beta \right\rfloor \cdot \gamma + \left\lfloor\frac{\Delta y_t - y_{min}}{y_{max} - y_{min}}\gamma \right\rfloor \ .
\end{split}
\label{eq: final_output}
\end{equation}
Finally, we process the past \( T \) states \( { \{(\theta_t, x_t, y_t, \mathbf{I}_t ) \}}_{t=1}^{T} \) and tokenize them into a discrete sequence \( {\{ (\phi_{t}, v_{t}, \mathbf{q}_{t}) \}}_{t=1}^{T} \). Specifically, $v_t$ denotes the tokenized representation of the vehicle’s 2D location (discretized $x$ and $y$ coordinates).

\begin{figure*}[t]
    \centering
    \includegraphics[width=0.9\textwidth]{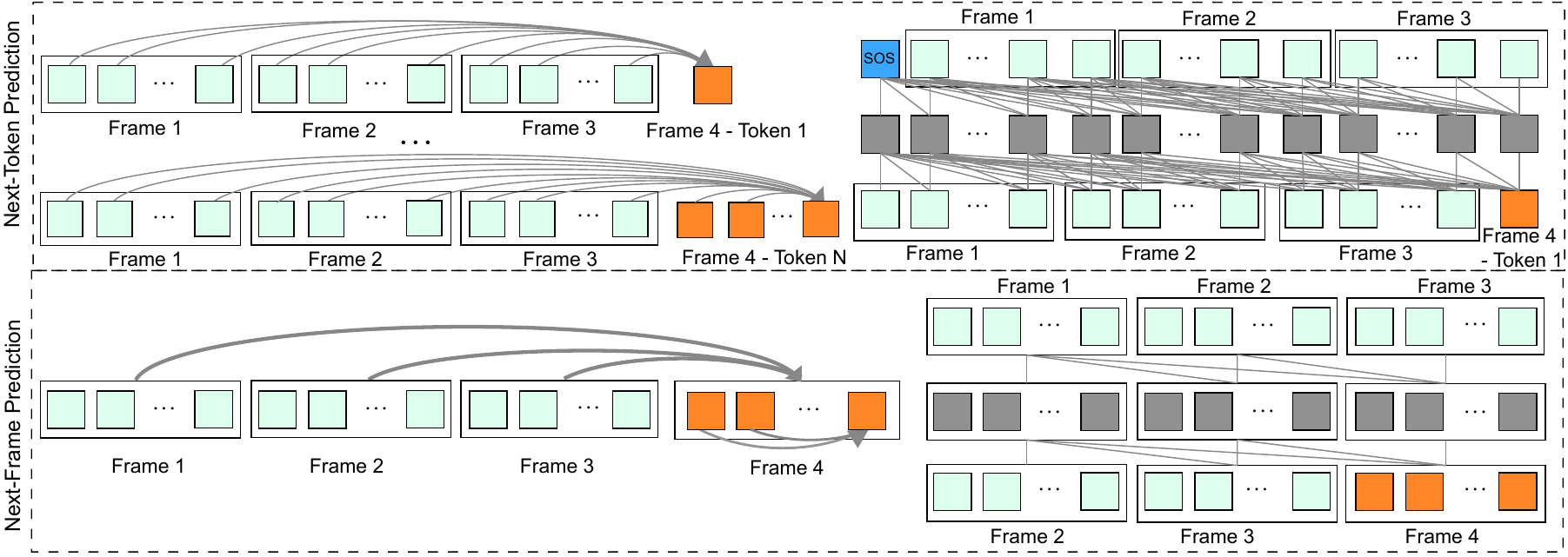}
    \caption{Illustration of next-token prediction and next-frame prediction. Unlike next-token prediction, next-frame prediction significantly reduces attention operations, decreasing computational load and alleviating attention dilution.}
    \label{fig: gpt cmp}
\end{figure*}

\subsection{Prelimilaries of Next-Token Prediction}
While future state prediction can be viewed as a long-context \textit{next-token prediction} task, directly applying language AR models~\cite{radford2018improving, radford2019language} to video generation faces key bottlenecks (see Section I). To probe these limitations, we visualize the attention map from the fifth Transformer block of a naive AR model~\cite{radford2019language}, as shown in Figure~\ref{fig:attn_map_vis}. For clarity, we use 16 image tokens, 1 control token, and consider a history of T=5 past states. Two notable phenomena can be observed: 1) \textbf{Control Signal Degradation}: Image tokens in the query tend to attend predominantly to the abundant image tokens in the key (highlighted by red boxes), while largely ignoring the control token that should also be referenced. This demonstrates that the influence of control signals is diluted in the attention mechanism. 2) \textbf{Temporal Disconnection}: Query tokens tend to attend mainly to key tokens from temporally nearby states (indicated by brighter colors toward the right), while neglecting tokens from more distant timesteps. This hinders long-term dependency learning and degrades performance.

\subsection{World Model}
\label{sec:world_model}
To mitigate attention dilution, enhance temporal connections, and reduce redundant computations, we propose a novel \textbf{next-frame AR module} for autonomous driving. Specifically, we decouple the long-sequence attention into two parts: one for temporal information attention and the other for multimodal information attention. The temporal information attention allows each token to focus solely on its past information, while the multimodal information attention enables each token to attend to all relevant information within the current time step, as shown in Figure~\ref{fig: gpt cmp}. This operation decouples the sequence, enabling separate handling of temporal and multimodal information. By doing so, it mitigates attention dilution, strengthens temporal connections, and significantly reduces the computational burden.
The input is historical $T$ states \( {\{ (\phi_{t}, v_{t}, \mathbf{q}_{t}) \}}_{t=1}^{T} \).
The goal is to predict the next state  $(\phi_{T+1}, v_{T+1}, \mathbf{q}_{T+1})$. 

\begin{figure}[t]
    \centering
    \begin{subfigure}{0.5\textwidth} 
        \centering
        \raisebox{10pt}{\includegraphics[width=\linewidth]{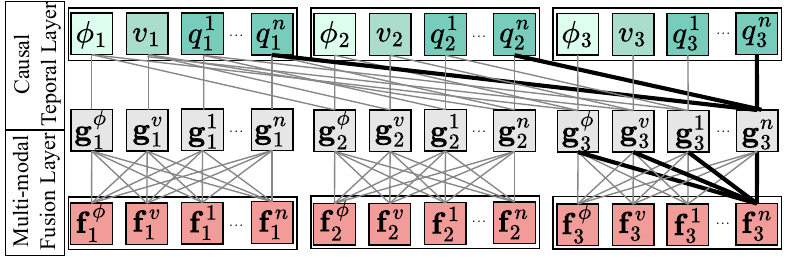}}
        \caption{Next-frame AR module}
        \label{fig:next_frame}
    \end{subfigure}
    \hfill %
    \begin{subfigure}{0.45\textwidth}
        \centering
        \includegraphics[width=\linewidth]{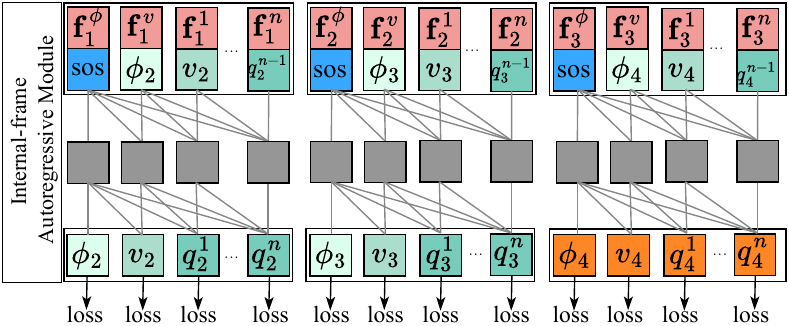} %
        \caption{Internal-frame AR module}
        \label{fig:inter_frame}
    \end{subfigure}
    \caption{Illustration of next-frame AR module and internal-frame AR module. In next-frame AR module (a), causal temporal layer links temporal dependencies at a location; multi-modal fusion layer integrates spatial information within a timestep. Internal-frame AR module (b) processes tokens in parallel within each timestep, preserving sequential dependencies inside the frame.}
\end{figure}

\begin{figure}[t]
    \centering
    \includegraphics[width=0.9\textwidth]{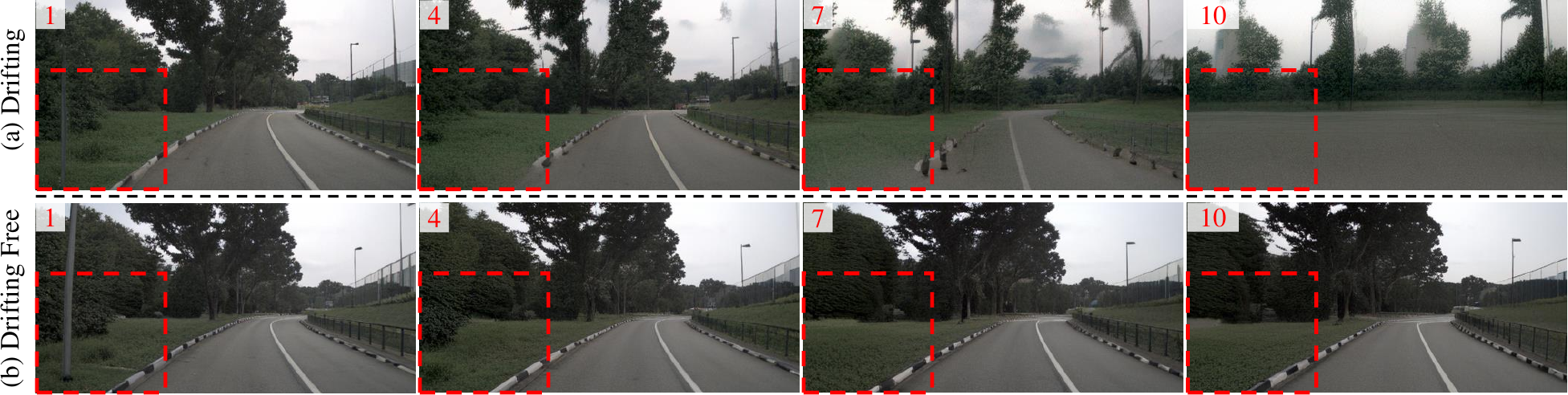}
    \caption{\textbf{Effectiveness of our random masking strategy}. Our masking strategy effectively mitigates AR drifting. Without it, the world model suffers severe drift, causing generated videos to degrade rapidly after the 10th frame.}
    \label{fig: autoregressive drift}
\end{figure}

\begin{table*}[t]
\label{tab:fvd}
\centering
\caption{Comparisons on the NuScenes~\cite{caesar2020nuscenes}. \colorbox{best2}{Blue}: NuScenes seen in training. \colorbox{best}{Orange}: zero-shot.
} 
\resizebox{\textwidth}{!}{\begin{tabular}{lccccc|ccc}
\toprule
\multirow{2}{*}{\textbf{Metric}}
 & \cellcolor{best2}DriveDreamer & \cellcolor{best2}WoVoGen & \cellcolor{best2}Drive-WM & \cellcolor{best2}Vista & \cellcolor{best2}\NickName &\cellcolor{best}DriveGAN & \cellcolor{best}GenAD (OpenDV) & \cellcolor{best}\NickName \\
 & \cellcolor{best2}\cite{wang2023drivedreamer} & \cellcolor{best2}\cite{lu2023wovogen} & \cellcolor{best2}\cite{wang2023driving}  & \cellcolor{best2}\cite{gao2024vista} & \cellcolor{best2}Ours & \cellcolor{best}\cite{santana2016learning}  & \cellcolor{best}\cite{yang2024generalized}  &  \cellcolor{best}Ours\\
\midrule
\textbf{FID $\downarrow$}  & \cellcolor{best2}52.6 & \cellcolor{best2}27.6 & \cellcolor{best2}15.8  & \cellcolor{best2}\underline{6.9} & \cellcolor{best2}\textbf{6.5} & \cellcolor{best}73.4  & \cellcolor{best}15.4  & \cellcolor{best}7.4\\
\textbf{FVD $\downarrow$}  & \cellcolor{best2}452.0 & \cellcolor{best2}417.7 & \cellcolor{best2}122.7  & \cellcolor{best2}\underline{89.4}&  \cellcolor{best2}\textbf{86.0}& \cellcolor{best}502.3  & \cellcolor{best}184.0 &
\cellcolor{best}90.9\\
\textbf{Max Duration}  & \cellcolor{best2}4s & \cellcolor{best2}2.5s & \cellcolor{best2}8s  &\cellcolor{best2} \underline{15s} & \cellcolor{best2}\textbf{40s} & \cellcolor{best}N/A  & \cellcolor{best} 4s  & \cellcolor{best} \textbf{40s} \\
\textbf{Max Frames}  & \cellcolor{best2}48 & \cellcolor{best2} 5 & \cellcolor{best2}16   &\cellcolor{best2} \underline{150} & \cellcolor{best2}\textbf{400} & \cellcolor{best}N/A  & \cellcolor{best} 8  & \cellcolor{best} \textbf{400} \\
\bottomrule
\end{tabular}
}
\label{result:compare1}
\end{table*}

\noindent
\textbf{Next-Frame AR Module.} The structure is shown in Figure~\ref{fig:next_frame}. It is composed of a causal temporal layer and a multimodal fusion layer. This decouples the processing of temporal and multimodal information, which significantly reduces the length of the sequence in attention. Specifically, given past $T$ states \( {\{ (\phi_{t}, v_{t}, \mathbf{q}_{t}) \}}_{t=1}^{T} \), we unfold the 2D image tokens into a 1D form in zig-zag order to get \( {\{ (\phi_{t}, v_{t}, q^1_{t}, ..., q_{t}^{n}) \}}_{t=1}^{T}, n = H\times W \). The causal temporal layer in Figure~\ref{fig:next_frame} captures dependencies across different time steps at the same spatial location.

\begin{equation}
\begin{aligned}
    \mathbf{g}_{t}^{\phi} &= \mathcal{F}_a(q^\phi_{1}, ..., q^\phi_{t}),\ t \in [1, T], \\
    \mathbf{g}_{t}^{v} &= \mathcal{F}_a(q^v_{1}, ..., q^v_{t}),\ t \in [1, T], \\
    \mathbf{g}_{t}^{j} &= \mathcal{F}_a(q^j_{1}, ..., q^j_{t}),\ t \in [1, T], j \in [1, n].
\end{aligned}
\end{equation}

The multi-modal fusion layer in Figure~\ref{fig:next_frame} focuses on fusing information across different spatial locations at the same time step:
\begin{equation}
    \{\mathbf{f}_{t}^{\phi}, \mathbf{f}_{t}^{v}, \mathbf{f}_{t}^{1}, ..., \mathbf{f}_{t}^{n}\} = \mathcal{F}_b(\mathbf{g}^\phi_{t}, \mathbf{g}^{v}_{t}, \mathbf{g}_t^{1}, ..., \mathbf{g}_{t}^{n}),\ t \in [1, T].
\end{equation}

However, we find that only using next-frame AR module yields suboptimal results. Recently, multiple image-generation works~\cite{sun2024autoregressive} propose that an AR pipeline for next-token prediction generates better images, and even outperforms diffusion methods. Inspired by this, we propose an \textbf{internal-frame AR module} to enhance image quality while also improving the model's controllability.

\noindent\textbf{Internal-Frame AR Module.}
As shown in Figure~\ref{fig:inter_frame}, given temporal-multimodal output features $\{\mathbf{f}_{t}^{\phi}, \mathbf{f}_{t}^{v}, \mathbf{f}_{t}^{1}, ..., \mathbf{f}_{t}^{n}\}$, we add them with prefix internal-state tokens $\{[sos], \phi_{T+1}, v_{T+1}, q_{T+1}^{1}, ..., q_{T+1}^{n-1}\}$ to predict next token:
\begin{equation}
q^{j}_{T+1} = \mathcal{F}_c([sos]+\mathbf{f}^{\phi}_{T}, ..., q^{j-1}_{T+1}+\mathbf{f}^{j}_{T}), j \in [1, n].
\end{equation}

The internal-frame AR module performs autoregression within each single time step, processing tokens in parallel to improve efficiency while still maintaining sequential dependencies within the frame.
However, each front view image is discretized into $512$ tokens, while only $2$ tokens represent the pose (orientation and location). 
This imbalance can cause the model to overlook pose signals, leading to unsatisfactory controllable generation.
To address this, we propose a novel \textbf{balanced attention} that adds regularization to achieve more precise control by prioritizing ego state tokens in the attention mechanism, rather than attending to all tokens equally. 
Specifically, we have \( m \) modalities with token counts $\{ n_1, ..., n_m\}$. If position \( i \) belongs to modality \( j \), the balanced weight $\hat{z}_i$:
\begin{equation}
    \begin{aligned}
        \hat{z}_{i} = z_{i} + \frac{1}{m n_j}, \texttt{softmax}(\hat{z}_i) = \frac{e^{\hat{z}_i}}{\sum_{k} e^{\hat{z}_k}}, 
    \end{aligned}
\end{equation}
where $\texttt{softmax}(\hat{z}_i)$ is the final weight. 

\noindent\textbf{Random Masking Strategy.} 
Teacher forcing~\cite{williams1989learning} is widely used in current next-token prediction models. Models use past ground-truth tokens to predict the immediate next token. 
However, current next-token models struggle with stability when handling continuous data. For instance, when generating a video autoregressively—unlike text~\cite{brown2020language}—small errors in frame-to-frame predictions gradually accumulate beyond the training horizon, causing the model to diverge.
To address this, we propose a random masking strategy (RMS), where some tokens from ground-truth tokens are randomly dropped out. There is a $\epsilon$ probability that the sequence performs random masking, and a $(1-\epsilon)$ probability that all ground-truth tokens are used. Each masked token has a $\eta$ probability of being randomly replaced with an arbitrary value.

We use cross-entropy loss for training:
\begin{equation}
    \mathcal{L}_{WM} = - \sum_{t=1}^{T+1} \sum_{i=1}^{n} \log p(q^i_{t} | \mathbf{q}_{<t}, q^{j<i}_{t}, \phi_{\leq t}, v_{\leq t}).
\end{equation}

\subsection{Decoder}
\label{sec:decoder}
The next-state tokens $(\phi_{T+1}, v_{T+1}, \mathbf{q}_{T+1})$ are predicted using the world model. Then we can leverage the decoder to generate the corresponding relative orientation $\Delta \theta_{T+1}$, relative location $(\Delta x_{T+1}, \Delta y_{T+1})$, and the reconstructed image $\mathbf{I}_{T+1}$for that state. This process allows us to map the predicted latent representations back into physical outputs, including both spatial and visual data.

\noindent
\textbf{Video Decoder.} For the predicted image tokens $\mathbf{q}_{T+1}$, we retrieve the corresponding features from the codebook in video tokenizer. The feature is fed into a causal attention layer and several convolution layers to get the video frames.

\noindent
\textbf{Vehicle Pose Decoder.} For the predicted relative orientation token \(\phi_{T+1}\) and relative location token \(v_{T+1}\), the corresponding values are obtained via the inverse of Eq.~\ref{eq: final_output} as follows:
\begin{equation}
\begin{split}
    \Delta \theta_t &= \theta_{min} + \frac{\phi_t}{\alpha} \left(\theta_{max} - \theta_{min}\right), \\
    \Delta x_t &= x_{min} + \frac{1}{\beta}\left\lfloor\frac{v_t}{\gamma}\right\rfloor \left(x_{max} - x_{min}\right), \\
    \Delta y_t &= y_{min} + \left(\frac{v_t}{\gamma}v_t - \left\lfloor\frac{v_t}{\gamma}\right\rfloor\right) \left(y_{max} - y_{min}\right).    
\end{split}
\end{equation}
\begin{figure}[t]
    \centering
    \includegraphics[width=0.95\linewidth]{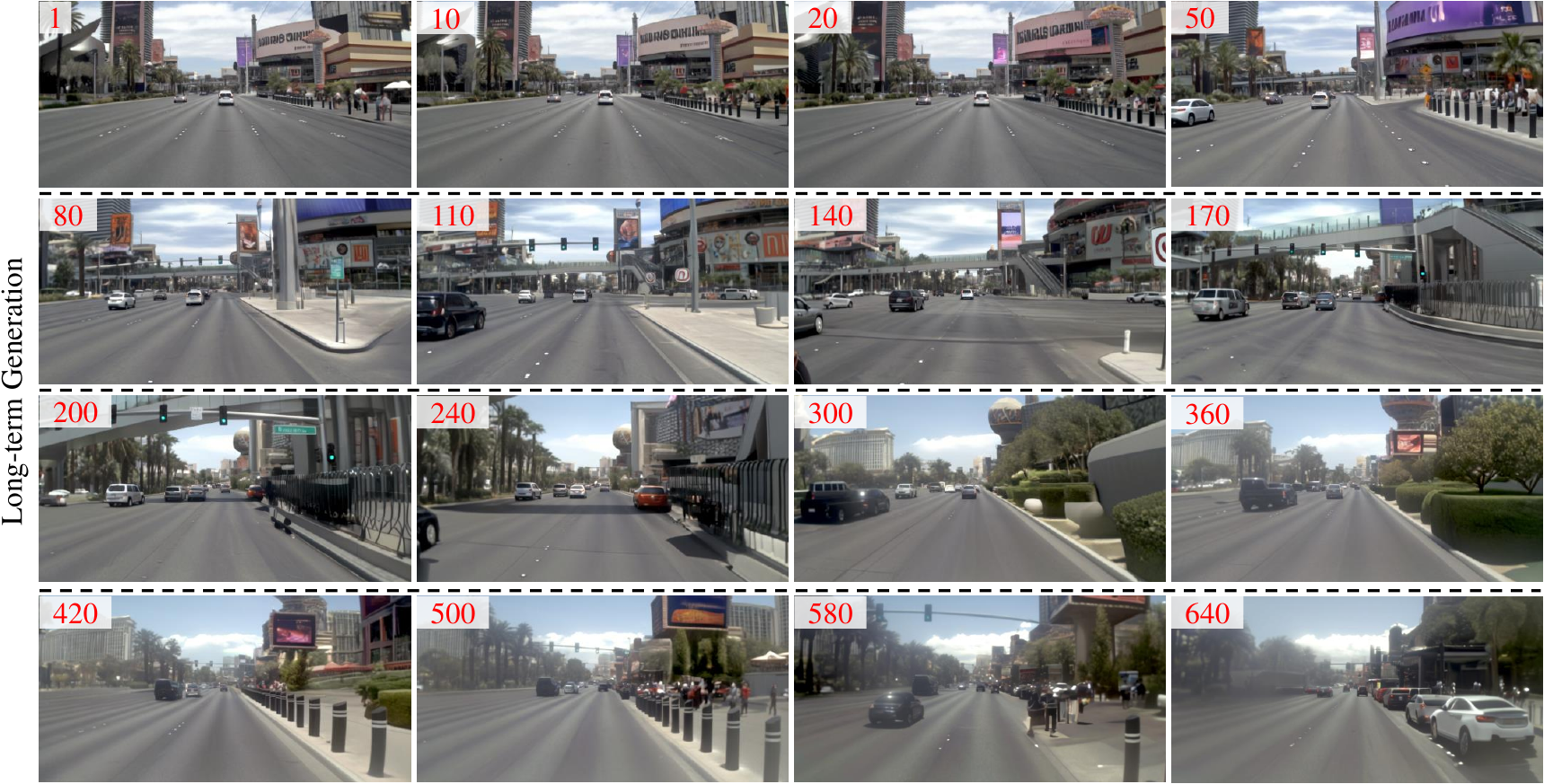}
    \caption{Long duration video generation. We present some videos generated by our method, each with 640 frames at 5Hz, i.e. 128 seconds. Please notice the coherent 3D scene structures captured by our method across different frames.}
    \label{fig:long_term}
\end{figure}
\section{Experiments}
\subsection{Implementation Details}

\noindent\textbf{Tokenizer and Decoder.} The video tokenizer has $70M$ parameters. The size of adopted codebook is set to $16,384$.
The model is trained for $1,000K$ steps with a total batch size of $128$ on $32$ NVIDIA 4090 GPUs, using images from Openimages~\cite{kuznetsova2020open}, COCO~\cite{lin2014microsoft}, YoutubeDV~\cite{zhang2022learning}, and NuPlan~\cite{caesar2021nuplan} datasets.
We train the temporal-aware VQVAE using a combination of three loss functions: charbonnier loss~\cite{Charbonnier_loss}, perceptual loss~\cite{zhang2018unreasonable}, and codebook loss~\cite{van2017neural}. 

\noindent\textbf{World Model.} The world model has $1B$ parameters and is trained on video sequences.
The model is conditioned on $15$ frames to predict the next frame. 
The values of $\epsilon$ and $\eta$ in RMS are set to 0.5 and 0.3, respectively.
It is trained on over $3456$ hours of human driving data. 120 hours come from the publicly available NuPlan~\cite{caesar2021nuplan} dataset, and 3336 hours consist of private data.
Training is conducted over $12$ days, spanning $450K$ iterations with a batch size of $64$, distributed across $64$ NVIDIA A100 GPUs. 
The model is fine-tuned on NuScenes~\cite{caesar2020nuscenes} for one day.

\begin{figure}[t]
    \centering
    \includegraphics[width=0.4\linewidth]{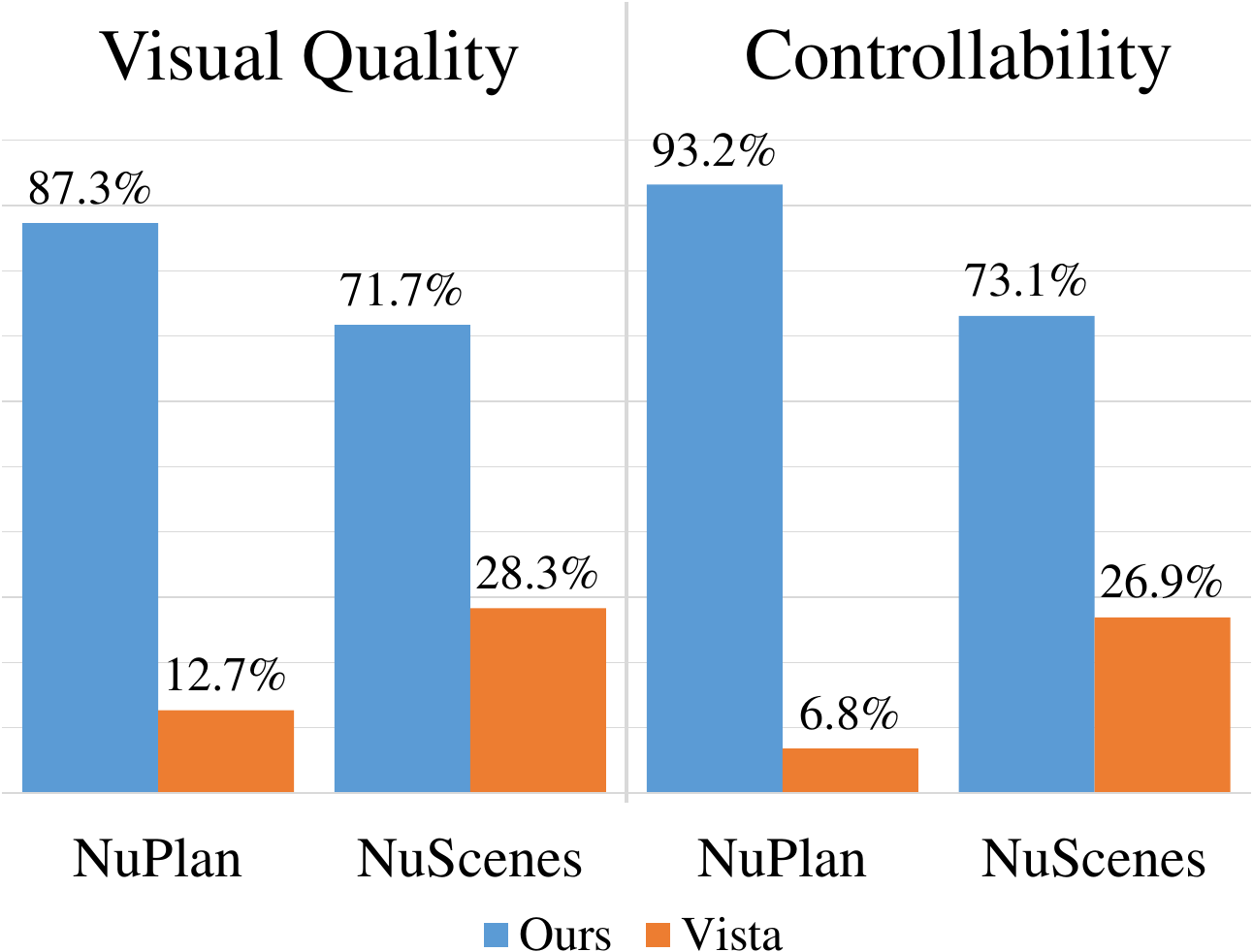}
    \caption{User study. The value represents the percentage of times one model is favored over the other. Our model surpasses Vista~\cite{gao2024vista} across all metrics.}
    \label{fig:user_study}
\end{figure}

\noindent
\textbf{Evaluation Dataset and Metrics.} We use $200$ video clips from the NuPlan~\cite{caesar2021nuplan} test dataset as our test set. Additionally, we include $150$ video clips from the NuScenes~\cite{caesar2020nuscenes} test set as part of our evaluation following Vista~\cite{gao2024vista}. The quality of video generation is assessed using the Frechet Video Distance (FVD), and we also report the Frechet Inception Distance (FID) to evaluate image generation quality. All generated videos used for FVD and FID evaluation have the same frame length, following the setting in Vista~\cite{gao2024vista}.

\subsection{Comparison of Generalization}
We provide quantitative comparisons on the NuScenes~\cite{caesar2020nuscenes} dataset in Table~\ref{result:compare1}. \NickName~trained with NuScenes achieves the best performance. \NickName~trained without it reaches performance comparable to Vista. Additionally, our model can generate the longest video sequences—up to 40 seconds. We also adopt human evaluation for more reliable analysis~\cite{wang2024lavie}. As shown in Figure~\ref{fig:user_study}, our model not only generates higher visual quality but also effectively adapts to control inputs compared with the SoTA model Vista~\cite{gao2024vista}.

\subsection{Results of Long-term Video Generation}
One of the key advantages of our method is its ability to generate long-duration videos. As shown in Figure~\ref{fig:long_term}, we visualize one long-duration video generated by our model. By conditioning on just $15$ frames, our model can generate up to $640$ future frames at $10$ Hz, resulting in $64$-second videos with strong temporal consistency. These results demonstrate that our model maintains high video fidelity and preserves 3D structural integrity across the generated frames.
In contrast, previous methods often struggle with drifting or degradation in long-duration videos. The ability to generate extended video sequences underscores our model's potential for tasks that require long-term predictions, such as autonomous driving or video synthesis in complex dynamic environments.

\begin{table}[t]
\centering
\caption{Quantitative comparison of VQVAE methods on NuPlan~\cite{caesar2021nuplan}.}
\resizebox{0.65\linewidth}{!}{
\begin{tabular}{l|cccc} 
\toprule
VQVAE Methods & {FVD$_{12}$ ↓} & {FID ↓} & {PSNR ↑} & {LPIPS ↓} \\ 
\hline
VAR~\cite{tian2024visual}  & 164.66 & 11.75 & 22.35 & 0.2018 \\
VQGAN~\cite{esser2021taming}  & 156.58 & 8.46 & 21.52 & 0.2602 \\
Llama-Gen~\cite{sun2024autoregressive}  & 57.78 & 5.99 & 22.31 & 0.2054 \\
Llama-Gen~\cite{sun2024autoregressive} Finetuned  & 20.33 & 5.19 & 22.71  & 0.1909 \\
Temporal-aware (Ours)  & \textbf{14.66} & \textbf{4.29} & \textbf{23.82}  & \textbf{0.1828} \\

\bottomrule
\end{tabular}
}
\label{tab:vqvae_table}
\end{table}

\begin{table}[t]
\setlength{\tabcolsep}{0.01\linewidth}
\caption{End-to-end motion planning performance on the NuScenes~\cite{caesar2020nuscenes} dataset. $^*$ represents only using the front camera as input.}
\centering
\resizebox{0.95\linewidth}{!}{
\begin{tabular}{l|lc|cccc|cccc}
\toprule
\multirow{2}{*}{Method} & \multirow{2}{*}{Input} & \multirow{2}{*}{Auxiliary Supervision} &
\multicolumn{4}{c|}{L2 (m) $\downarrow$} & 
\multicolumn{4}{c}{Collision Rate (\%) $\downarrow$} \\
&& & 1s & 2s & 3s & \cellcolor{gray!30}Avg. & 1s & 2s & 3s & \cellcolor{gray!30}Avg.  \\
\midrule
ST-P3~\cite{hu2022stp3} & Camera & Map \& Box \& Depth & 1.33 & 2.11 & 2.90 & \cellcolor{gray!30}2.11 & 0.23 & 0.62 & 1.27 & \cellcolor{gray!30}0.71  \\
UniAD~\cite{uniad} & Camera & { \footnotesize Map \& Box \& Motion \& Tracklets \& Occ}  & {0.48} & {0.96} & {1.65} & \cellcolor{gray!30}{1.03} & {0.05} & {\textbf{0.17}} & \textbf{0.71} & \cellcolor{gray!30}{\textbf{0.31}}  \\
OccNet~\cite{tong2023scene} & Camera & 3D-Occ \& Map \& Box & 1.29 & 2.13 & 2.99 & \cellcolor{gray!30}2.14 & 0.21 & 0.59 & 1.37 & \cellcolor{gray!30}0.72  \\
OccWorld~\cite{zheng2024occworld} & Camera & 3D-Occ & 0.52 & 1.27 & 2.41 & \cellcolor{gray!30}1.40 & 0.12 & 0.40 & 2.08 & \cellcolor{gray!30}0.87  \\
VAD-Tiny~\cite{vad}  & Camera & Map \& Box \& Motion  & 0.60 & 1.23 & 2.06 & \cellcolor{gray!30}1.30 & 0.31 & 0.53 & 1.33 & \cellcolor{gray!30}0.72  \\
VAD-Base~\cite{vad} & Camera & Map \& Box \& Motion & 0.54 & 1.15 & 1.98 & \cellcolor{gray!30}1.22 & 0.04 & 0.39 & 1.17 & \cellcolor{gray!30}0.53 \\
GenAD~\cite{zheng2024genad} & Camera & Map \& Box \& Motion & {\textbf{0.36}} & {\textbf{0.83}} & {\textbf{1.55}} & \cellcolor{gray!30}{\textbf{0.91}} & 0.06 & {0.23} & {1.00} & \cellcolor{gray!30}{0.43} \\
\midrule
\textbf{\textbf{Ours}} & Camera$^*$ & None & 0.68 & 1.20 & 1.90 & \cellcolor{gray!30}1.26 & \textbf{0.03} & 0.22 & 0.90 & \cellcolor{gray!30}0.38 \\
\bottomrule
\end{tabular}%
}
\label{tab:plan_nusc}
\end{table}

\begin{figure}[t]
    \centering
    \includegraphics[width=0.45\linewidth]{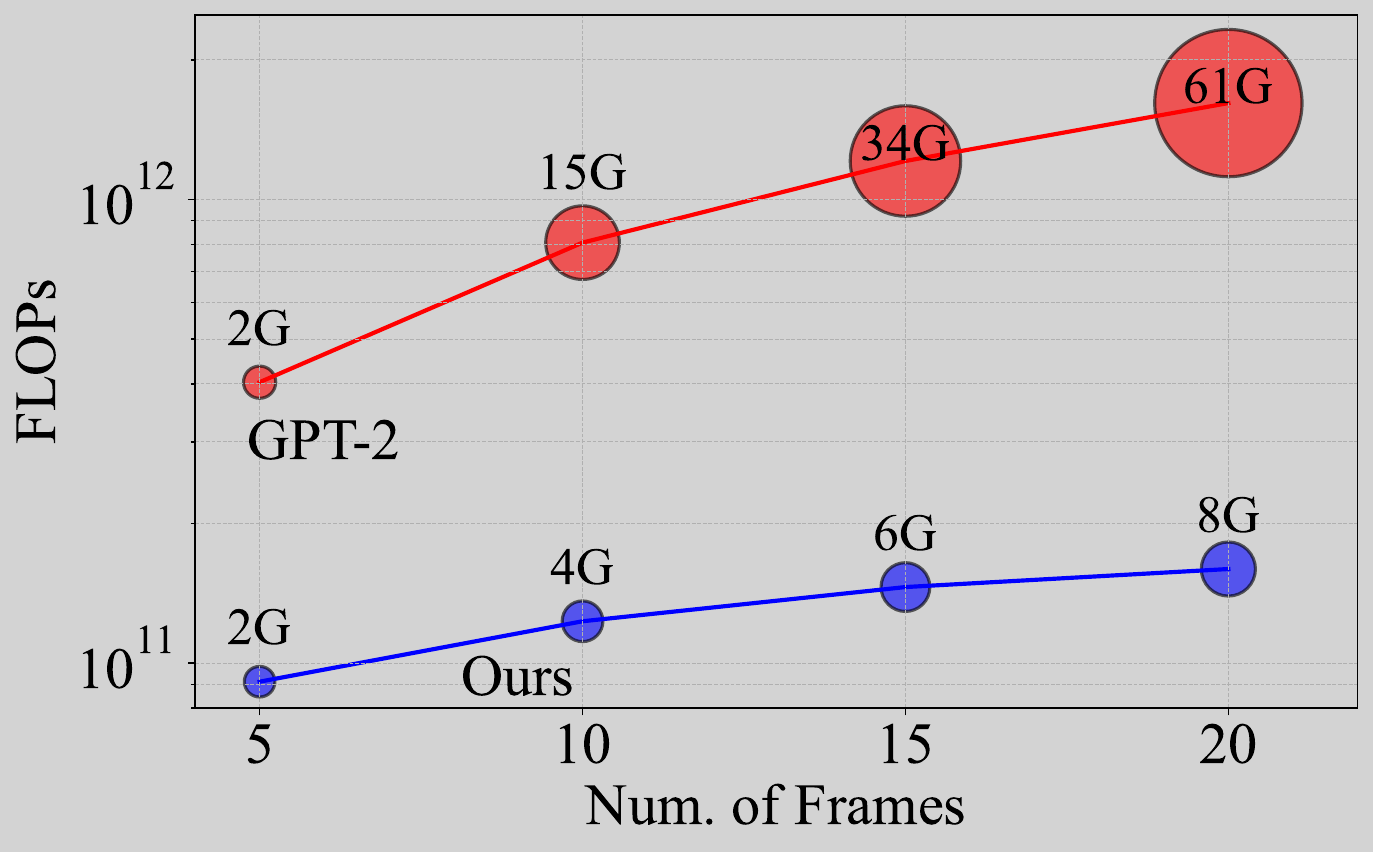}
    \caption{Comparison of GPT-2 and \NickName~(ours).}
    \label{fig:flops}
\end{figure}

\begin{table}[t]
\caption{Ablation study of Balanced Attention (BA) and Random Masking Strategy (RMS) on NuPlan~\cite{caesar2021nuplan}.}
\centering
\resizebox{0.5\linewidth}{!}{
\begin{tabular}{lccc} 
\toprule
Methods   & FVD$_{10}$ ↓ & FVD$_{25}$ ↓ & FVD$_{40}$ ↓ \\
\midrule
\textit{w/o} BA \& \textit{w/o} RMS         & 463.21 & 603.88 & 665.15 \\
\textit{w/o} BA                & 240.25 & 319.35 & 361.94 \\
\textit{w/o} RMS               & 214.69 & 283.71 & 324.58 \\
\textbf{Ours}    & \textbf{122.37} &  \textbf{144.26} & \textbf{159.46} \\
\bottomrule
\end{tabular}
}
\label{tab:ablation_3}
\end{table}
    
\subsection{Quantitative Comparison of Image Tokenizers} 

We further evaluate our temporal-aware image tokenizer against those proposed in other works. Since the image tokenizer is part of a VQVAE, we assess the encoding-decoding performance of these VQVAEs instead. The experiments, conducted on the NuPlan~\cite{caesar2021nuplan} dataset, are summarized in Table~\ref{tab:vqvae_table}.
The VQVAE models from VAR~\cite{tian2024visual} and VQGAN~\cite{esser2021taming} demonstrate reasonable image quality in terms of PSNR and LPIPS scores, but both fall short on FID and FVD metrics. In contrast, Llama-Gen's VQVAE~\cite{sun2024autoregressive} shows significant improvements in FID and FVD scores. After fine-tuning it on driving scenes, we observe further enhancements in FVD performance.
Ultimately, our temporal-aware VQVAE outperforms all others, enhancing temporal consistency and achieving the best scores across all four metrics.

\subsection{Ablation Study}
\noindent
\textbf{Comparison of BA and RMS.} As shown in Table~\ref{tab:ablation_3}, removing either Balanced Attention (BA) or Random Masking Strategy (RMS) degrades performance, while using both together achieves the best results. This indicates that BA and RMS are complementary: BA helps preserve control information, and RMS improves long-term temporal modeling, leading to significantly lower FVD.

\noindent
\textbf{Discussion on Next-Token Prediction.} 
We compare the memory usage and FLOPs of our \NickName~ with the next-token prediction architecture, specifically GPT-2~\cite{radford2019language}, which processes tokens sequentially across all frames. As shown in Figure~\ref{fig:flops}, GPT-2's memory usage grows rapidly with sequence length, making it inefficient for long sequences.
In contrast, our method decouples temporal and multimodal dependencies for more efficient computation.

\subsection{Comparison of Trajectory Planning}
To demonstrate the applicability of our framework in autonomous driving, we evaluate its vehicle control capability through trajectory planning on the NuScenes benchmark~\cite{caesar2020nuscenes}, following prior works~\cite{uniad, hu2022stp3} with L2 error and collision rate as the evaluation metrics, as shown in Table~\ref{tab:plan_nusc}. Requiring only front-view images, our method delivers competitive SOTA-level performance and robustness with high practicality for perception-limited scenarios.
\section{Conclusion and Future Work}

In conclusion, \NickName~addressed two key limitations of previous video generation models in autonomous driving by leveraging a next-token prediction framework to produce longer, high-fidelity video predictions without control signal degradation or computational inefficiency. Unlike traditional methods that struggled with coherence in long sequences, \NickName~generated realistic, structured video sequences while enabling precise action control.
Our proposed next-frame AR module adopted a next-frame prediction strategy to model temporal coherence between consecutive frames and spatial reasoning within frames. We also implemented an internal-frame AR module to capture spatial information in each frame.
Future work will integrate multimodal and multi-view data to improve understanding and control, advancing autonomous driving capabilities.

%
%
%

\bibliographystyle{splncs04}
\bibliography{main}
\end{document}